\documentclass[letterpaper, 10 pt, conference]{ieeeconf}  

\IEEEoverridecommandlockouts                              

\usepackage{scalerel}
\usepackage{graphicx}
\graphicspath{{./results/images/}}
\usepackage{amsmath} 
\usepackage{amssymb}  
\usepackage[dvipsnames]{xcolor}
\usepackage[short,nocomma]{optidef}
\DeclareMathOperator*{\argmax}{arg\,max}

\title{\LARGE \bf
Searching for waveforms on spatially-filtered epileptic ECoG
}

\author{Carlos H. Mendoza-Cardenas$^{1}$ and Austin J. Brockmeier$^{2}$
\thanks{$^{1}$Carlos H. Mendoza-Cardenas is with the Department of Electrical \& Computer Engineering, University of Delaware, Newark, Delaware, USA. He has been partly funded by Minciencias (Colombia). {\tt\small cmendoza@udel.edu}}%
\thanks{$^{2}$Austin J. Brockmeier is with the Department of Electrical \& Computer Engineering and the Department of Computer \& Information Sciences,  University of Delaware, Newark, Delaware, USA. {\tt\small ajbrock@udel.edu}}%
}

\usepackage{hyperref}
\begin{document}
	
This is the accepted version of the paper to appear in the 2021 10th International IEEE/EMBS Conference on Neural Engineering \href{neuro.embs.org}{(NER)}.\\
	
\textsuperscript{\textcopyright} 2021 IEEE. Personal use of this material is permitted. Permission from IEEE must be obtained for all other uses, in any current or future media, including
reprinting/republishing this material for advertising or promotional purposes, creating new
collective works, for resale or redistribution to servers or lists, or reuse of any copyrighted
component of this work in other works.

\maketitle
\thispagestyle{empty}
\pagestyle{empty}

\begin{abstract}

Seizures are one of the defining symptoms in patients with epilepsy, and due to their unannounced occurrence, they can pose a severe risk for the individual that suffers it. New research efforts are showing a promising future for the prediction and preemption of imminent seizures, and with those efforts, a vast and diverse set of features have been proposed for seizure prediction algorithms. However, the data-driven discovery of nonsinusoidal waveforms for seizure prediction is lacking in the literature, which is in stark contrast with recent works that show the close connection between the waveform morphology of neural oscillations and the physiology and pathophysiology of the brain, and especially its use in effectively discriminating between normal and abnormal oscillations in electrocorticographic (ECoG) recordings of epileptic patients. Here, we explore a scalable, energy-guided waveform search strategy on spatially-projected continuous multi-day ECoG data sets. Our work shows that data-driven waveform learning methods have the potential to not only contribute features with predictive power for seizure prediction, but also to facilitate the discovery of oscillatory patterns that could contribute to our understanding of the pathophysiology and etiology of seizures.

\end{abstract}

\section{INTRODUCTION}


The spontaneous and usually unforeseen nature of seizures is a major risk factor in individuals with epilepsy, causing unintentional injuries, drowning, anxiety, depression, and, overall, a premature mortality rate up to three times higher than the general population \cite{WorldHealthOrganization2019}. Furthermore, around one-third of the population with epilepsy has a drug-resistant epilepsy. All this has motivated research efforts in seizure prediction for more than three decades, with recent advances in methods and devices showing a promising future for the prediction and preemption of impending seizures \cite{Kuhlmann2018}.

A great diversity of electroencephalographic (EEG) features have been developed and used in seizure prediction algorithms, from correlation \cite{Williamson2012} and phase synchronization \cite{Mormann2000} features, to chaos measures \cite{Iasemidis2005} (see \cite{Mormann2007} and \cite{Gadhoumi2016} for a review of EEG features and algorithms in seizure prediction). Most of the measures developed so far are obtained through either a linear or non-linear transformation, or a space/time-delay embedding, of the EEG time series, without regard to the waveform morphologies of the neural oscillations.

Recent works have highlighted the importance of stereotyped nonsinusoidal EEG waveform morphologies in brain physiology and pathophysiology \cite{Gerber2016,Cole2017,Cole2017a}, and in particular, its utility to distinguish between normal and abnormal high-frequency oscillations (HFO) in electrocorticographic (ECoG) recordings from epileptic patients \cite{Liu2018}. Although several works have developed shift-invariant, and data-driven methods that learn waveforms from EEG data in an unsupervised manner \cite{Brockmeier2016,Jas2017,Dupre2018}, only one work, to the best of our knowledge, has explored the discovery and use of stereotyped temporal waveforms in seizure prediction \cite{Cui2018}. Motivated by these previous works on waveform learning and its almost absent application to seizure detection, we present here an exploratory study of a waveform search strategy applied to continuous multi-day ECoG recordings from two epileptic patients.

In seizure prediction algorithms, it is customary to define two classes of segments in an epileptic EEG recording: (1) preictal segments that can last from minutes to hours, located just before the onset of a seizure, usually with a time gap called \textit{minimum intervention time} \cite{Mormann2007}; and (2) interictal segments well-separated in time from any seizure. In this work, we aim to find waveforms identified by their spatial pattern that are discriminative of the interictal and preictal state of an ECoG recording. Discriminative in the sense that the waveforms are more prevalent in one state compared to the other. To guide our search for those waveforms, we first look at specific spectral bands. We use the common spatial patterns (CSP) method to compute spatial filters that maximize the spectral band power of ECoG windows from one condition while minimizing it for windows from the other condition.

In contrast to our work, the bag-of-waves (BoWav) representation proposed in \cite{Cui2018} uses two codebooks of preictal and interictal waveforms that are built by applying the $k$-means clustering algorithm on a random sample of small, single-channel windows of a multi-channel EEG recording, without any type of channel re-referencing or spatial filtering. The BoWav features are histograms of the codebook waveforms and are built by sliding a window through all the EEG channels during a given period.

With a scalable spatial projection (CSP) and our energy-guided strategy, we found waveforms that are discriminative of the preictal and interictal state, and, in some cases, are well-known epileptiform patterns. Our work offers preliminary evidence of the potential that data-driven waveform learning methods have in automating the discovery of EEG patterns, designing new features from those patterns that could be used for seizure prediction, and helping advance our understanding of the pathophysiology of the brain.

Finally, in addition to the waveform search strategy explored here, we performed an extensive visual inspection of the 28 data sets of epileptic ECoG data that were used in \cite{Kini2019} and built a preprocessing pipeline after identifying some artifacts that were common among several of those continuous multi-day ECoG recordings. We work here with two of those data sets. Our code base, including our preprocessing pipeline, is publicly available at \verb|github.com/chmendoza/cspwave| for research reproducibility and for the benefit of other researchers that might be interested in working with those data sets.


\section{Methods}

\subsection{Notation}

Let $g_\mathbf{w}(\mathbf{X})=\mathbf{w}^\text{T}\mathbf{X} \in \mathbb{R}^L$ denote the projection of $\mathbf{X} \in \mathbb{R}^{C \times L}$ onto the spatial filter $\mathbf{w} \in \mathbb{R}^C$, with $C$ and $L$ denoting the number of EEG channels and time points, respectively. We will call this product a CSP signal. Let $\mathcal{U} = \{\mathbf{u}_i \in \mathbb{R}^{L}, i \in [m]\}$ be a set of CSP signals, and $h_k(\mathcal{U}) = \argmax_{\mathcal{I}\subset [m], \vert\mathcal{I}\vert=k}\sum_{i \in \mathcal{I}} \lVert \mathbf{u}_i\rVert^2_2$ denote the set of indices of the $k$ CSP signals in $\mathcal{U}$ with the highest energy, with $[m] = \{1,...,m\}$, $\vert \mathcal{I} \vert$ being the cardinality of $\mathcal{I}$, and $\lVert \cdot \rVert_2$ denoting the Euclidean norm. 

\subsection{Data}

The data sets used in this work are continuous long-term multichannel ECoG recordings from two epileptic patients publicly available at \verb|ieeg.org| \cite{Wagenaar2013}. Table \ref{tab:data_info} presents some characteristics of the data after preprocessing: the patient's age in years; the number of channels ($C$) and seizures (N.S); the seizure type (S.T), either simple partial seizure (SPS) or complex partial seizure (CPS); and the total length of the preictal and interictal intervals in hours (h) and minutes (m). We discarded the same EEG channels per patient, and use the same seizure annotations as in \cite{Kini2019}; more details about the data can be found in that reference. The relevant clinical seizure markings for this work are the earliest EEG change (EEC), which is the point in time with the first clear and sustained change from the patient's EEG baseline before the seizure onset, and the end of the seizure. We assume that the activity happening between those two time points represents the ictal state. Following the guidelines of two Kaggle competitions of seizure prediction \cite{AES2014,AES2016}, we define the interictal state as the activity that happens at least four hours away from the ictal state, and the preictal state as the activity that occurs in the 1-hour interval that goes from 1:05 to 0:05 before EEC, with five minutes of minimum intervention time.

\begin{table}[b]
	\caption{Data characteristics}
	\label{tab:data_info}
	\centering
	\begin{tabular}{lrrrlll}
		\textbf{Name} & \textbf{Age} & \textbf{\textit{C}} & \textbf{N.S.} & \textbf{S.T.} & \textbf{Preictal} & \textbf{Interictal}  \\ 
		\hline \\[-1.5ex]
		HUP070 & 33 &  63    & 3             & SPS           & 2h 6m     &  40h 32m     \\
		HUP078 & 54 &  101   & 3             & CPS           & 1h 30m     & 45h 23m    
	\end{tabular}
\end{table}

\subsection{Preprocessing}

\begin{figure}[t]
    \includegraphics[width=\linewidth]{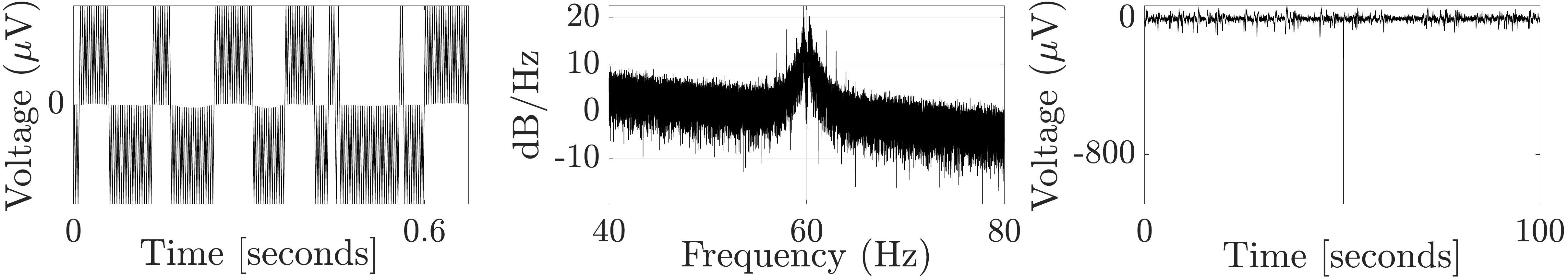}
	\caption{Example artifacts. (left) Rail-to-rail oscillation, (middle) Broad 60 Hz peak after applying \textit{cleanLineNoise} and (right) Anomalous spike.}
	\label{fig:artifact-examples}
\end{figure}

After an extensive visual inspection of the raw data, we identified several artifacts (see Fig. \ref{fig:artifact-examples}) and applied the following preprocessing pipeline. We discarded segments with missing samples, or that have a constant amplitude for more than 25 ms. Due to length restrictions imposed by some spectral filtering functions used in later preprocessing stages, we discarded segments that were less than 5 seconds long. Some segments had a rail-to-rail oscillation, consisting of frequent peaks that were saturated at the same amplitude. Since the repeated occurrence of a specific amplitude value is unlikely for time-varying signals that are quantized at a 24-bit rate, we dealt with that artifact by discarding a segment if at least one of its amplitude values has a relative frequency higher than 5\%.  We applied a \mbox{1-Hz} high pass filter \cite{Delorme2004}, and then the \textit{cleanLineNoise} function \cite{Bigdely-Shamlo2015} at \{60, 120, 180\} Hz to remove the power line artifact. After this, we noticed that the spectrum of some segments had a broad and strong peak at 60 Hz, different from the narrow peak that was eliminated after applying \textit{cleanLineNoise}, likely due to the modulation of the 60 Hz power line signal by movement artifacts. We addressed that distortion by rejecting a segment if its power in the 45-55 Hz band is lower than its power in the 55-65 Hz band. We also found that some segments had anomalous spikes in the time-series that were abnormally high in amplitude and slope. We set an \textit{ad hoc} threshold of 70 $\mu$V in the change of amplitude between consecutive samples to detect and remove those anomalous spikes; we removed the 2-minute windows centered at those spikes. Finally, the data from all the patients were resampled to 512 Hz.

\subsection{Spatial filters}

Let $\mathcal{X}_s = \{\mathbf{X}_{i,s} \in \mathbb{R}^{C\times L}\}^{N_s}_{i=1}$ be a training set of EEG non-overlapping windows, sampled at random and uniformly across all the EEG segments from either the preictal ($s=1$) or interictal ($s=2$) condition, where $C$, $L$ and $N_s$ are the number of channels, time points and windows, respectively. To further restrict and guide our search for the most discriminative morphologies on the spatially-projected space, we pass each EEG window through a band-pass filter $f_\text{B}(\cdot)$, with passband $\mathrm{B} \in \mathcal{B}=\{\delta[1.5-4], \theta[4-8], \alpha[8-15], \beta_{low}[15-26], \beta_{high}[26-35], \gamma_{low}[35-50], \gamma_{mid}[50-74], \gamma_{high}[76-120], \mathrm{HFO}[120-220]\}$, in units of Hz. We use the method of common spatial patterns (CSP) \cite{Blankertz2008} to find two spatial filters in $\mathbb{R}^{C}$, $\mathbf{w}_1$ and $\mathbf{w}_2$, that maximize the energy of the CSP signal under the preictal and interictal state, respectively. $\mathbf{w}_1$ can be expressed as the solution to the following Rayleigh quotient maximization problem

\begin{argmaxi}
	{\mathbf{w}}{\frac{\mathbf{w}^\text{T}\mathbf{\Sigma}_1\mathbf{w}}{\mathbf{w}^\text{T}(\mathbf{\Sigma}_1+\mathbf{\Sigma}_2)\mathbf{w}},}
	{\label{eq:rayleigh}}{\mathbf{w}_1=}
\end{argmaxi}

%
%
\noindent with 
\begin{equation}
\mathbf{\Sigma}_s = \frac{1}{N_s(L-1)}\sum_{i=1}^{N_s} f_\text{B}(\mathbf{X}_{i,s})f_\text{B}(\mathbf{X}_{i,s})^\text{T}
\end{equation}

\noindent being the average estimated covariance of the set of $N_s$ band-pass filtered windows from condition $s \in \{1,2\}$. Analogously, Eq. \eqref{eq:rayleigh} can be solved to obtain $\mathbf{w}_2$, by using $\mathbf{\Sigma}_2$ on the numerator. Since the covariance matrices, and the spatial filters, are computed on the temporally filtered windows, we obtain nine sets of two spatial filters, one set per spectral band.

\subsection{Waveform search strategy}

Our strategy to search for waveforms on the preictal and interictal EEG segments has two main steps: 1) identify the time windows with the highest energy after temporal and spatial filtering, and 2) spatially filter the EEG data located in those time windows, without temporal filtering. More formally, let $\mathcal{V}_s=\{\mathbf{V}_{i,s} \in \mathbb{R}^{C\times L}\}^{M_s}_{i=1}$ be a test set of $M_s$ non-overlapping windows of length $L$ from condition $s \in \{1,2\}$, sampled in the same way as $\mathcal{X}_s$. We filter those windows in time and space to get $\tilde{\mathcal{U}}_{s,t} = \{\tilde{\mathbf{u}}_{i,s,t} \in \mathbb{R}^L: \tilde{\mathbf{u}}_{i,s,t}=g_{\mathbf{w}_t}\left(f_\text{B}(\mathbf{V}_{i,s})\right), i \in [M_s]\}$, for $s,t \in \{1,2\}$. We then find the index set of the $k$ CSP signals (or time windows) with the highest energy in $\tilde{\mathcal{U}}_{s,t}$, denoted as $\mathcal{I}_{k,s,t} = h_k(\tilde{\mathcal{U}}_{s,t})$. Finally, we apply $\mathbf{w}_1$ and $\mathbf{w}_2$ to the EEG windows in $\mathcal{V}_s$ indexed by $\mathcal{I}_{k,s,t}$, thus getting the set of CSP signals $\mathcal{U}_{k,s,t} = \left\{\mathbf{u}_{j,s,t} \in \mathbb{R}^L: \mathbf{u}_{j,s,t} = g_{\mathbf{w}_t}\left(\mathbf{V}_{i_j,s}\right),\; i_j \in \mathcal{I}_{k,s,t},\; j \in [k] \right\}$, for $k \in \mathbb{N}$, and $s,t \in \{1,2\}$.



\section{Results}

We did an 80/20 split of the available data for training and testing. The number of windows for training was $N_1 = \{6000, 4320\}$ and $N_2 = \{116400, 130000\}$ for \{HUP070, HUP078\}, and $M_1=\{1500, 1080\}$ and $M_2=\{29100, 32500\}$ for testing. The window length was one second ($L=512$) for training and testing.

\begin{figure}[t]
	\mbox{}\\[0.7\baselineskip]
	\includegraphics[width=\linewidth]{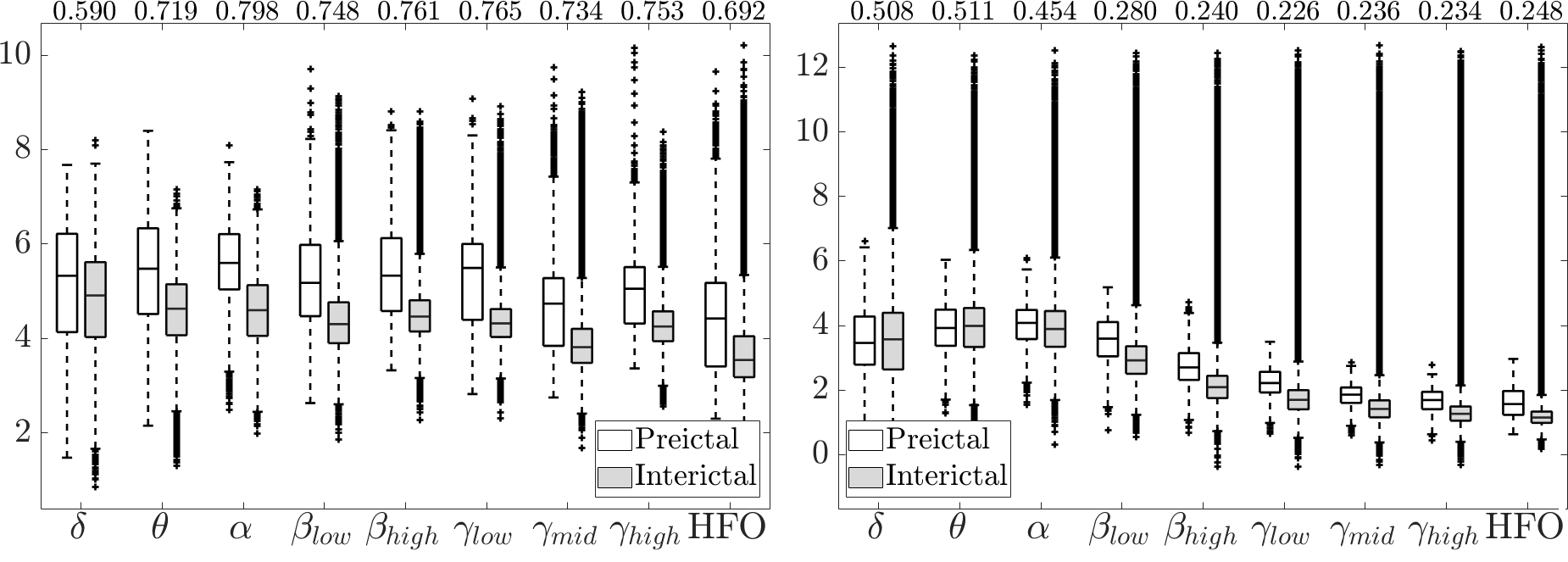}
	\caption{Log-energy ($\mu \mathrm{V}^2$) of CSP signals in HUP070. (left) Log-energy of signals in $\tilde{\mathcal{U}}_{s,1}$, for $s \in \{1,2\}$. (right) Log-energy of $\tilde{\mathcal{U}}_{s,2}$. Labels in the bottom axis are the spectral bands over which the CSP filter was optimized, and numbers in the top axis are the AUC values.\label{fig:boxplot_AUC_HUP070}}
	
	\mbox{}\\[1.5\baselineskip]
	\includegraphics[width=\linewidth]{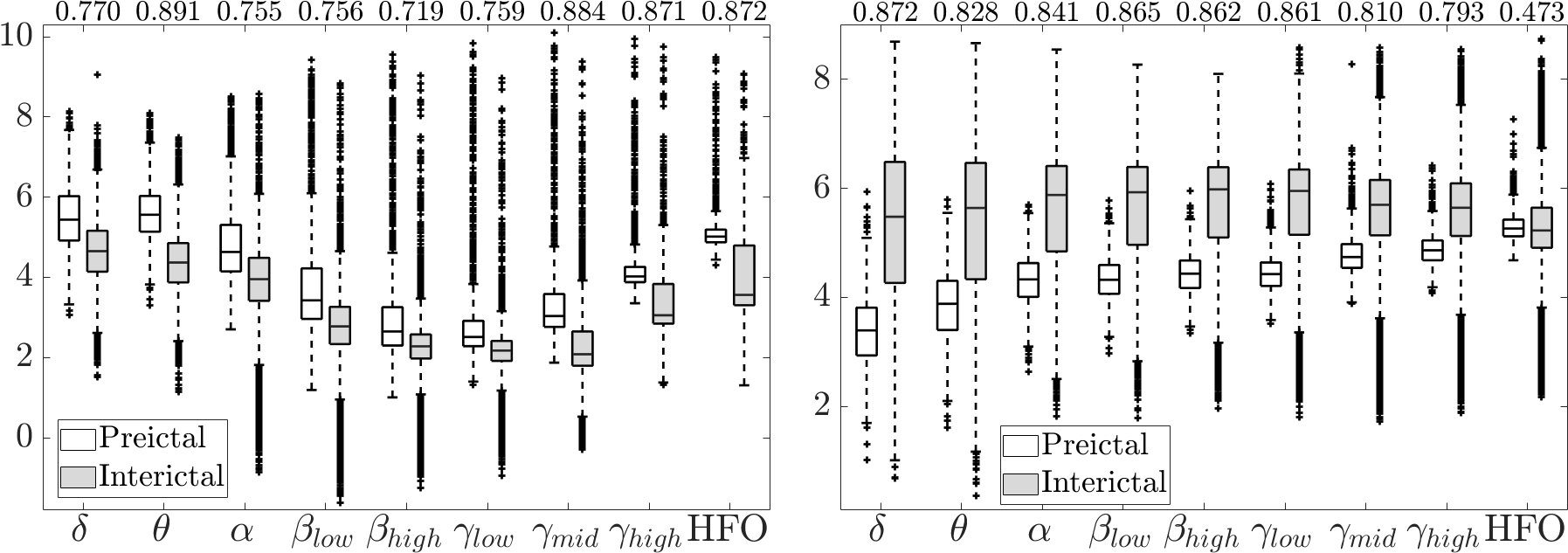}
	\caption{Log-energy ($\mu \mathrm{V}^2$) of CSP signals in HUP078. (left) Log-energy of signals in $\tilde{\mathcal{U}}_{s,1}$, for $s \in \{1,2\}$. (right) Log-energy of $\tilde{\mathcal{U}}_{s,2}$. Labels in the bottom axis are the spectral bands over which the CSP filter was optimized, and numbers in the top axis are the AUC values.\label{fig:boxplot_AUC_HUP078}}
	
	\mbox{}\\[1.5\baselineskip]	
		\includegraphics[width=\linewidth]{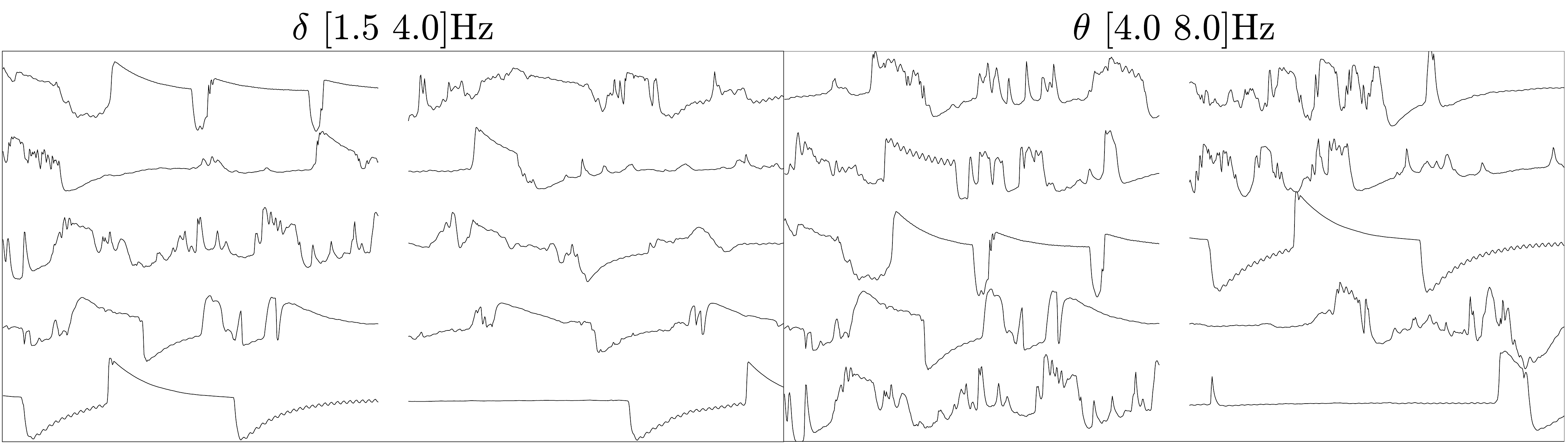}
	\caption{Signals in $\mathcal{U}_{10,2,2}$, for HUP070 and the two spectral bands with highest test AUC. \label{fig:topk_artifacts}}
	\mbox{}\\[-2\baselineskip]
\end{figure}

Although this work explores an energy-guided strategy to find waveforms that are discriminative of the interictal and preictal state, and not a classification method itself, we use the area under the receiver operating characteristic curve (AUC) of a binary classifier to quantify the performance of $\mathbf{w}_1$ and $\mathbf{w}_2$ in discriminating between the two states. The $i$-th EEG window $\mathbf{V}_{i,s}$ from state $s$ is filtered in space and time, yielding $\tilde{\mathbf{u}}_{i,s,t}$, and a binary classifier makes a prediction $\hat{s}$ of the state of that window using a hard threshold on the energy of $\tilde{\mathbf{u}}_{i,s,t}$, with $i \in [M_s]$ and $s, t, \hat{s} \in \{1,2\}$. Fig. \ref{fig:boxplot_AUC_HUP070} and Fig. \ref{fig:boxplot_AUC_HUP078} show a pair-wise comparison of the boxplot of the log-energy of each CSP filter output when that spatial filter is applied to temporally-filtered windows from both conditions, as well as the AUC values of such binary classifier. We found that HUP070 has large-energy artifacts that we did not account for in our preprocessing pipeline (see Fig. \ref{fig:topk_artifacts}), which caused poor performance of $\mathbf{w}_2$. In contrast, HUP078 did not have those artifacts, and thus the improved performance of $\mathbf{w}_2$ for that data set. The susceptibility of CSP to outliers is well-known \cite{Blankertz2008}, and a more robust variant of CSP that uses some type of regularization \cite{Wang2016}, a divergence-based framework that accounts for outliers \cite{Samek2014}, or an information-theoretic statistic \cite{Brockmeier2014}, could be explored to solve this problem.

\begin{figure}[t]


    \mbox{}\\[0.3\baselineskip]
		\includegraphics[width=\linewidth]{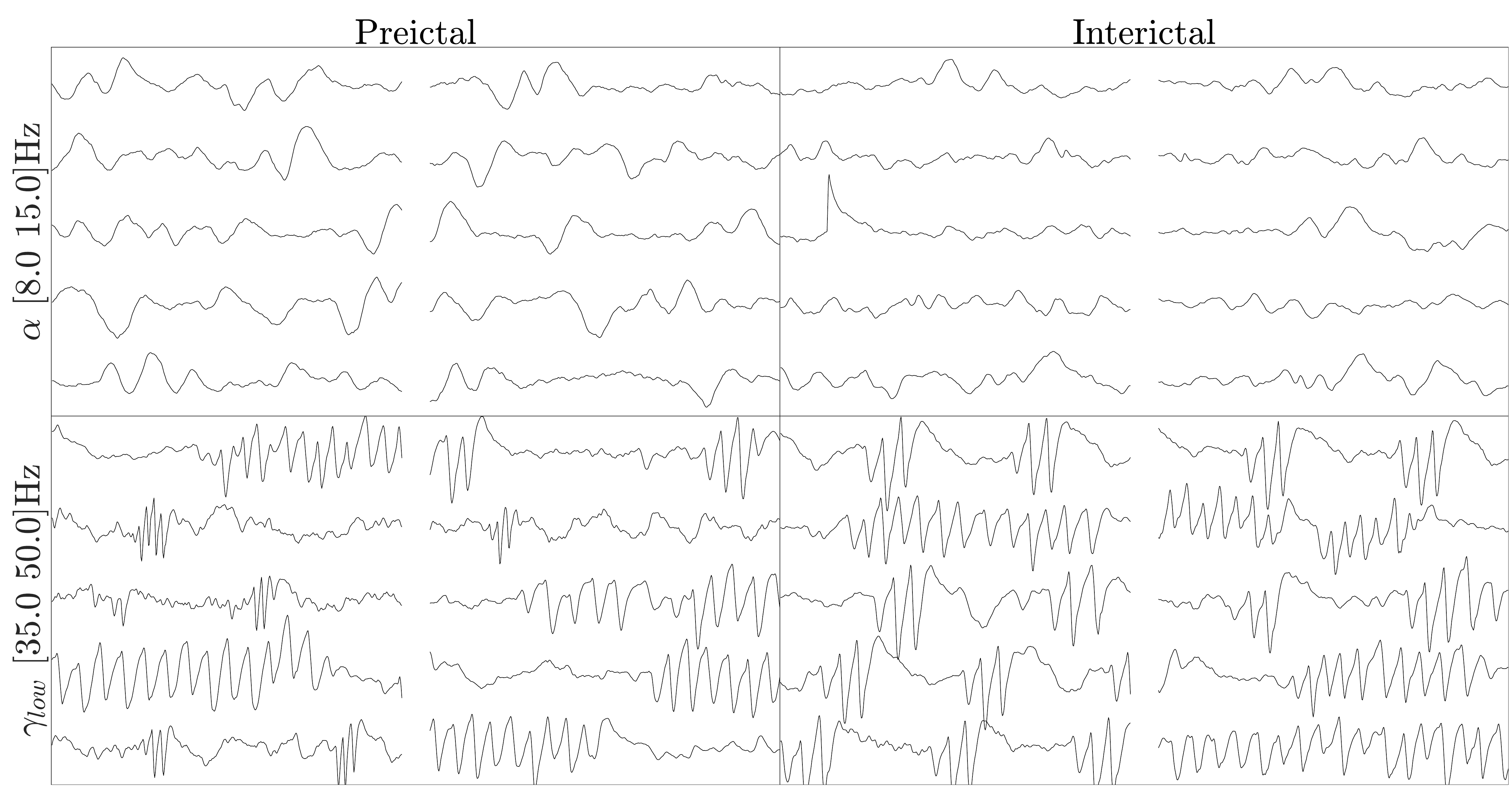}
	\caption{Top 10 CSP signals in HUP070 with highest energy from each condition, after applying the $\mathbf{w}_1$ optimized for a given spectral band. Signals in the top row are scaled by 2x to show waveform shape details.\label{fig:topk_waves_HUP070}}
	
	\mbox{}\\[\baselineskip]
	\includegraphics[width=\linewidth]{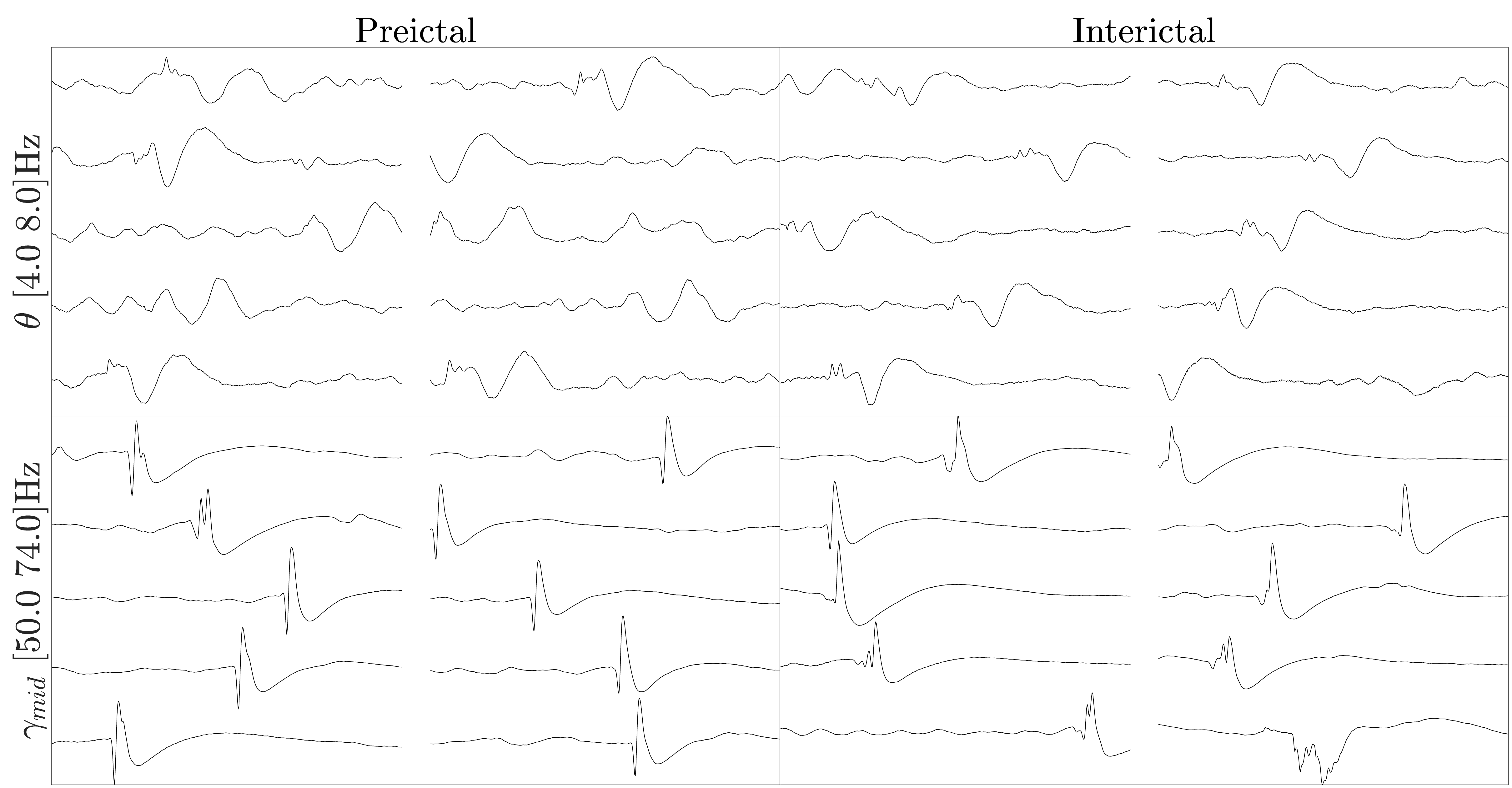}
	\caption{Top 10 CSP signals in HUP078 with highest energy from each condition, after applying the $\mathbf{w}_1$ optimized for a given spectral band. Signals in the top row are scaled by 10x to show waveform shape details.\label{fig:topk_waves_HUP078}}
	\mbox{}\\[-1.5\baselineskip]
\end{figure}

We used our waveform search strategy to visually explore the morphologies of the spatially-projected signals. Fig. \ref{fig:topk_waves_HUP070} and Fig. \ref{fig:topk_waves_HUP078} show the waveforms in $\mathcal{U}_{10,s,1}$ for the two datasets, the preictal ($s=1$) and interictal ($s=2$) conditions, and the two spectral bands with highest AUC in the test set. These waveforms are the result of applying $\mathbf{w}_1$ to the test EEG windows (in $\mathcal{V}_s$) corresponding to the 10 highest energy signals (in $\tilde{\mathcal{U}}_{s,1}$) after temporal and spatial filtering.

We found that using a scalable spatial filtering method (CSP) and our energy-guided search strategy, one can readily and automatically discover nonsinusoidal waveforms morphologies in the preictal and interictal state that are either qualitatively (visually) distinct or more prevalent in one condition. Some of those waveforms, like the sharp spikes in the bottom row of Fig. \ref{fig:topk_waves_HUP078}, are well-known epileptiform patterns \cite{Westmoreland1996}. Furthermore, we also found that at least one of the two spatial filters exhibits a high discriminative performance, measured by the AUC of a binary classifier. Our results, and the recent works on waveform morphology previously highlighted, suggest the potential that the data-driven discovery of stereotyped nonsinusoidal waveform shapes has, not only in the development of new features with predictive power for seizure prediction, but also in its contribution to our understanding of the pathophysiology and etiology of seizures.

\bibliographystyle{./IEEEtran}	
\bibliography{./NER21}

\begin{thebibliography}{10}
\providecommand{\url}[1]{#1}
\csname url@rmstyle\endcsname
\providecommand{\newblock}{\relax}
\providecommand{\bibinfo}[2]{#2}
\providecommand\BIBentrySTDinterwordspacing{\spaceskip=0pt\relax}
\providecommand\BIBentryALTinterwordstretchfactor{4}
\providecommand\BIBentryALTinterwordspacing{\spaceskip=\fontdimen2\font plus
\BIBentryALTinterwordstretchfactor\fontdimen3\font minus
  \fontdimen4\font\relax}
\providecommand\BIBforeignlanguage[2]{{%
\expandafter\ifx\csname l@#1\endcsname\relax
\typeout{** WARNING: IEEEtran.bst: No hyphenation pattern has been}%
\typeout{** loaded for the language `#1'. Using the pattern for}%
\typeout{** the default language instead.}%
\else
\language=\csname l@#1\endcsname
\fi
#2}}

\bibitem{WorldHealthOrganization2019}
\BIBentryALTinterwordspacing
{World Health Organization}, \emph{{Epilepsy: a public health imperative}},
  2019. [Online]. Available:
  \url{https://www.who.int/mental\_health/neurology/epilepsy/report\_2019/en/}
\BIBentrySTDinterwordspacing

\bibitem{Kuhlmann2018}
L.~Kuhlmann, K.~Lehnertz, M.~P. Richardson, B.~Schelter, and H.~P. Zaveri,
  ``{Seizure prediction - ready for a new era},'' \emph{Nat. Rev. Neurol.},
  vol.~14, no.~10, pp. 618--630, 2018.

\bibitem{Williamson2012}
\BIBentryALTinterwordspacing
J.~R. Williamson, D.~W. Bliss, D.~W. Browne, and J.~T. Narayanan, ``{Seizure
  prediction using EEG spatiotemporal correlation structure},'' \emph{Epilepsy
  Behav.}, vol.~25, no.~2, pp. 230--238, 2012. [Online]. Available:
  \url{http://dx.doi.org/10.1016/j.yebeh.2012.07.007}
\BIBentrySTDinterwordspacing

\bibitem{Mormann2000}
F.~Mormann, K.~Lehnertz, P.~David, and C.~{E. Elger}, ``{Mean phase coherence
  as a measure for phase synchronization and its application to the EEG of
  epilepsy patients},'' \emph{Phys. D Nonlinear Phenom.}, vol. 144, no.~3, pp.
  358--369, 2000.

\bibitem{Iasemidis2005}
L.~D. Iasemidis, D.~S. Shiau, P.~M. Pardalos, W.~Chaovalitwongse, K.~Narayanan,
  A.~Prasad, K.~Tsakalis, P.~R. Carney, and J.~C. Sackellares, ``{Long-term
  prospective on-line real-time seizure prediction},'' \emph{Clin.
  Neurophysiol.}, vol. 116, no.~3, pp. 532--544, 2005.

\bibitem{Mormann2007}
F.~Mormann, R.~G. Andrzejak, C.~E. Elger, and K.~Lehnertz, ``{Seizure
  prediction: The long and winding road},'' \emph{Brain}, vol. 130, no.~2, pp.
  314--333, 2007.

\bibitem{Gadhoumi2016}
\BIBentryALTinterwordspacing
K.~Gadhoumi, J.~M. Lina, F.~Mormann, and J.~Gotman, ``{Seizure prediction for
  therapeutic devices: A review},'' \emph{J. Neurosci. Methods}, vol. 260, no.
  029, pp. 270--282, 2016. [Online]. Available:
  \url{http://dx.doi.org/10.1016/j.jneumeth.2015.06.010}
\BIBentrySTDinterwordspacing

\bibitem{Gerber2016}
E.~M. Gerber, B.~Sadeh, A.~Ward, R.~T. Knight, and L.~Y. Deouell,
  ``{Non-sinusoidal activity can produce cross-frequency coupling in cortical
  signals in the absence of functional interaction between neural sources},''
  \emph{PLoS One}, vol.~11, no.~12, pp. 1--19, 2016.

\bibitem{Cole2017}
S.~R. Cole and B.~Voytek, ``{Brain Oscillations and the Importance of Waveform
  Shape},'' \emph{Trends Cogn. Sci.}, vol.~21, no.~2, pp. 137--149, 2017.

\bibitem{Cole2017a}
S.~R. Cole, R.~van~der Meij, E.~J. Peterson, C.~de~Hemptinne, P.~A. Starr, and
  B.~Voytek, ``{Nonsinusoidal beta oscillations reflect cortical
  pathophysiology in parkinson's disease},'' \emph{J. Neurosci.}, vol.~37,
  no.~18, pp. 4830--4840, 2017.

\bibitem{Liu2018}
S.~Liu, C.~Gurses, Z.~Sha, M.~M. Quach, A.~Sencer, N.~Bebek, D.~J. Curry,
  S.~Prabhu, S.~Tummala, T.~R. Henry, and N.~F. Ince, ``{Stereotyped
  high-frequency oscillations discriminate seizure onset zones and critical
  functional cortex in focal epilepsy},'' \emph{Brain}, vol. 141, no.~3, pp.
  713--730, 2018.

\bibitem{Brockmeier2016}
A.~J. Brockmeier and J.~C. Pr{\'{i}}ncipe, ``{Learning recurrent waveforms
  within EEGs},'' \emph{IEEE Trans. Biomed. Eng.}, vol.~63, no.~1, pp. 43--54,
  2016.

\bibitem{Jas2017}
M.~Jas, T.~D. {La Tour}, U.~\c{S}im\c{s}ekli, and A.~Gramfort, ``{Learning the
  morphology of brain signals using alpha-stable convolutional sparse
  coding},'' in \emph{Adv. Neural Inf. Process. Syst.}, 2017, p.~10.

\bibitem{Dupre2018}
T.~D. la~Tour, T.~Moreau, M.~Jas, and A.~Gramfort, ``{Multivariate
  convolutional sparse coding for electromagnetic brain signals},'' in
  \emph{Adv. Neural Inf. Process. Syst.}, 2018, p.~11.

\bibitem{Cui2018}
\BIBentryALTinterwordspacing
S.~Cui, L.~Duan, Y.~Qiao, and Y.~Xiao, ``{Learning EEG synchronization patterns
  for epileptic seizure prediction using bag-of-wave features},'' \emph{J.
  Ambient Intell. Humaniz. Comput.}, 2018. [Online]. Available:
  \url{http://dx.doi.org/10.1007/s12652-018-1000-3}
\BIBentrySTDinterwordspacing

\bibitem{Kini2019}
L.~G. Kini, J.~M. Bernabei, F.~Mikhail, P.~Hadar, P.~Shah, A.~N. Khambhati,
  K.~Oechsel, R.~Archer, J.~Boccanfuso, E.~Conrad, R.~T. Shinohara, J.~M.
  Stein, S.~Das, A.~Kheder, T.~H. Lucas, K.~A. Davis, D.~S. Bassett, and
  B.~Litt, ``{Virtual resection predicts surgical outcome for drug-resistant
  epilepsy},'' \emph{Brain}, vol. 142, no.~12, pp. 3892--3905, 2019.

\bibitem{Wagenaar2013}
J.~B. Wagenaar, B.~H. Brinkmann, Z.~Ives, G.~A. Worrell, and B.~Litt, ``{A
  multimodal platform for cloud-based collaborative research},'' \emph{Int.
  IEEE/EMBS Conf. Neural Eng. NER}, pp. 1386--1389, 2013.

\bibitem{AES2014}
\BIBentryALTinterwordspacing
``{American Epilepsy Society Seizure Prediction Challenge},'' 2014. [Online].
  Available: \url{https://www.kaggle.com/c/seizure-prediction}
\BIBentrySTDinterwordspacing

\bibitem{AES2016}
\BIBentryALTinterwordspacing
``{Melbourne University AES/MathWorks/NIH Seizure Prediction},'' 2016.
  [Online]. Available:
  \url{https://www.kaggle.com/c/melbourne-university-seizure-prediction}
\BIBentrySTDinterwordspacing

\bibitem{Delorme2004}
A.~Delorme and S.~Makeig, ``{EEGLAB: An open source toolbox for analysis of
  single-trial EEG dynamics including independent component analysis},''
  \emph{J. Neurosci. Methods}, vol. 134, no.~1, pp. 9--21, 2004.

\bibitem{Bigdely-Shamlo2015}
N.~Bigdely-Shamlo, T.~Mullen, C.~Kothe, K.~M. Su, and K.~A. Robbins, ``{The
  PREP pipeline: Standardized preprocessing for large-scale EEG analysis},''
  \emph{Front. Neuroinform.}, vol.~9, no. JUNE, pp. 1--19, 2015.

\bibitem{Blankertz2008}
B.~Blankertz, R.~Tomioka, S.~Lemm, M.~Kawanabe, and K.~R. M{\"{u}}ller,
  ``{Optimizing spatial filters for robust EEG single-trial analysis},''
  \emph{IEEE Signal Process. Mag.}, vol.~25, no.~1, pp. 41--56, 2008.

\bibitem{Wang2016}
H.~Wang and X.~Li, ``{Regularized Filters for L1-Norm-Based Common Spatial
  Patterns},'' \emph{IEEE Trans. Neural Syst. Rehabil. Eng.}, vol.~24, no.~2,
  pp. 201--211, 2016.

\bibitem{Samek2014}
W.~Samek, M.~Kawanabe, and K.~R. Muller, ``{Divergence-based framework for
  common spatial patterns algorithms},'' \emph{IEEE Rev. Biomed. Eng.}, vol.~7,
  pp. 50--72, 2014.

\bibitem{Brockmeier2014}
A.~J. Brockmeier, E.~Santanna, L.~G. Giraldo, and J.~C. Principe,
  ``{Projentropy: Using entropy to optimize spatial projections},''
  \emph{ICASSP, IEEE Int. Conf. Acoust. Speech Signal Process. - Proc.}, pp.
  4538--4542, 2014.

\bibitem{Westmoreland1996}
\BIBentryALTinterwordspacing
B.~F. Westmoreland, ``{Epileptiform electroencephalographic patterns},''
  \emph{Mayo Clin. Proc.}, vol.~71, no.~5, pp. 501--511, 1996. [Online].
  Available: \url{http://dx.doi.org/10.4065/71.5.501}
\BIBentrySTDinterwordspacing

\end{thebibliography}

\end{document}